\newif\ifIEEEFORMAT
\ifdefined\BUILDIEEE
    \IEEEFORMATtrue
\else
    \IEEEFORMATfalse
\fi

\ifIEEEFORMAT
\documentclass[lettersize,journal]{IEEEtran}
\else
\documentclass[manuscript,screen,preprint]{acmart}
\fi

\ifIEEEFORMAT
\providecommand{\Description}[1]{}
\else
\acmJournal{TOMM}
\setcopyright{none}
\fi
\usepackage{qtree}
\usepackage{graphicx}
\usepackage{listings}
\usepackage{xcolor}
\usepackage{amsfonts}
\usepackage{mathrsfs}
\usepackage{amsmath}
\usepackage{amsthm}
\usepackage{amscd}
\usepackage{tabularx}
\usepackage{multirow}
\usepackage{float}
\usepackage{subcaption}
\usepackage{array}
\usepackage{algorithm}
\usepackage{bm}
\usepackage{comment}
\usepackage{makecell}
\usepackage{caption}

\usepackage{algpseudocode}
\algnewcommand\algorithmicforeach{\textbf{for each}}
\algdef{S}[FOR]{ForEach}[1]{\algorithmicforeach\ #1\ \algorithmicdo}

\title{PPEDCRF: Dynamic-CRF-Guided Selective Perturbation for Background-Based Location Privacy in Video Sequences}
\ifIEEEFORMAT
\author{Bo Ma, Weiqi Yan, Jinsong Wu}
\fi
\author{Bo Ma}
\email{rcn4743@aut.ac.nz}
\affiliation{%
  \institution{Resideo Technologies Inc.}
  \city{Scottsdale}
  \state{Arizona}
  \country{USA}
}
\affiliation{%
  \institution{Auckland University of Technology}
  \city{Auckland}
  \country{New Zealand}
}

\author{Weiqi Yan}
\affiliation{%
  \institution{Auckland University of Technology}
  \city{Auckland}
  \country{New Zealand}
}

\author{Jinsong Wu}
\authornote{Senior Member, IEEE}
\affiliation{%
  \institution{Guilin University of Electronic Technology}
  \city{Guilin}
  \country{China}
}
\ifIEEEFORMAT
\else
\keywords{location privacy, selective perturbation, dynamic conditional random field, visual place recognition, privacy--utility trade-off, video anonymization}
\ccsdesc[500]{Security and privacy~Privacy protections}
\ccsdesc[300]{Security and privacy~Privacy-preserving protocols}
\ccsdesc[300]{Computing methodologies~Computer vision tasks}
\ccsdesc[300]{Information systems~Multimedia information systems}
\fi

\begin{document}
\begin{abstract}
We propose PPEDCRF, a calibrated selective perturbation framework that protects \emph{background-based location privacy} in released video frames against gallery-based retrieval attackers. Even after GPS metadata are stripped, an adversary can geolocate a frame by matching its background visual cues to geo-tagged reference imagery; PPEDCRF mitigates this threat by estimating location-sensitive background regions with a dynamic conditional random field (DCRF), rescaling perturbation strength with a normalized control penalty (NCP), and injecting Gaussian noise only inside the inferred regions via a DP-style calibration rule.

On a controlled paired-scene retrieval benchmark with eight attacker backbones and three noise seeds, PPEDCRF reduces ResNet18 Top-1 retrieval accuracy from 0.667 to $0.361\pm0.127$ at $\sigma_0=8$ while preserving 36.14\,dB PSNR---an ${\approx}6$\,dB quality advantage over global Gaussian noise. Transfer across the eight-backbone seed-averaged benchmark is broadly supportive (23 of 24 backbone--gallery cells show negative~$\Delta$), while appendix-scale confirmation identifies MixVPR as a remaining adverse-transfer exception. Matched-operating-point analysis shows that PPEDCRF and global Gaussian noise converge in Top-1 privacy at equal utility, so the practical benefit is spatially concentrated perturbation that preserves higher visual quality at any given noise scale rather than stronger matched-utility privacy.
\vspace{1em}
\noindent\textbf{Code:} \url{https://github.com/mabo1215/PPEDCRF}
\end{abstract}
\maketitle

\section{Introduction}
\begin{figure*}[!t]
  \centering
  \includegraphics[width=\textwidth]{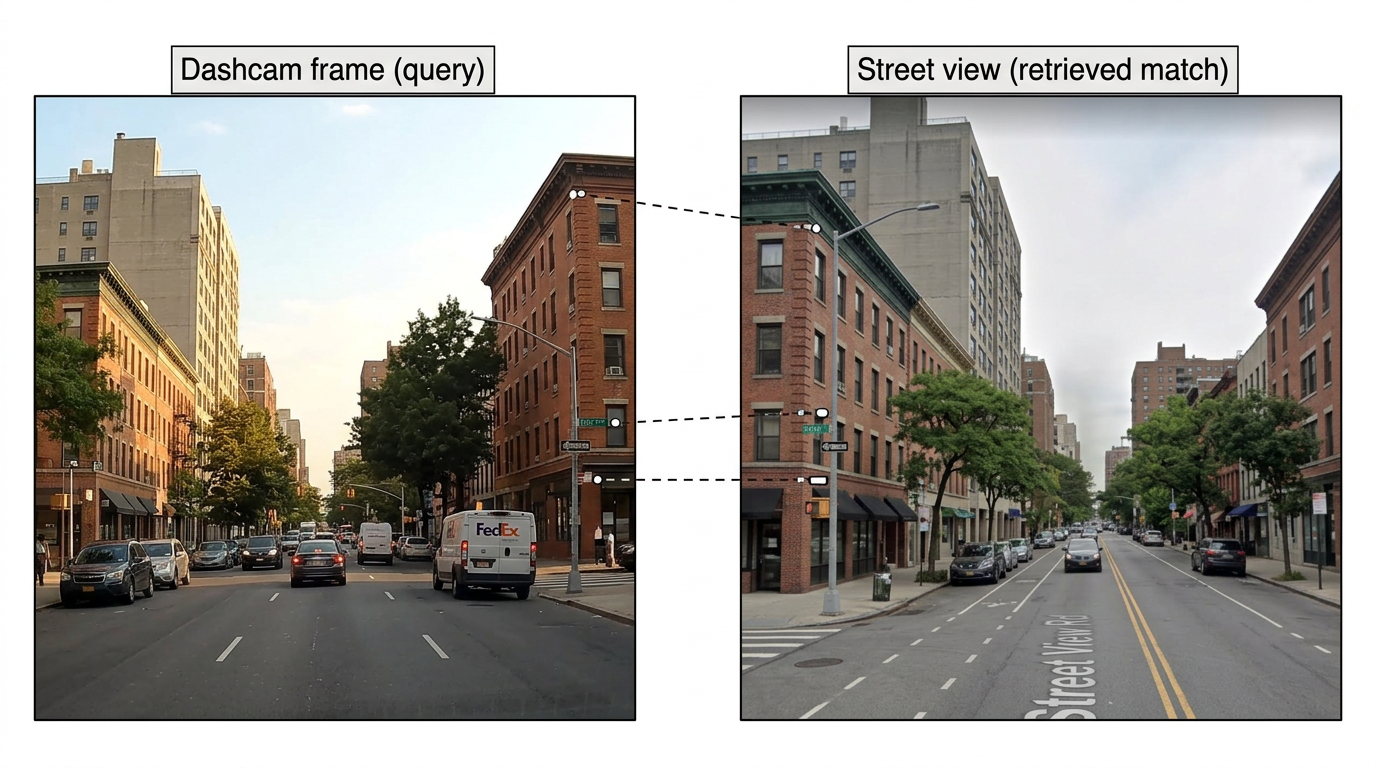}
  \caption{A released video frame can be matched to a geo-tagged reference image through background cues even when explicit metadata are removed.}
  \Description{A two-panel example showing a released video frame and a visually similar geo-tagged reference image, illustrating that background structures alone can reveal location.}
  \label{fig:twopic}
\end{figure*}

Mobile-recorded videos---from dashcams, visual-based ADAS platforms, bodycams, drones, and wearable cameras---are routinely uploaded for safety investigation, incident auditing, and data-driven model improvement. Removing explicit metadata such as GPS coordinates is necessary but not sufficient for privacy protection: the background of a single frame may contain distinctive visual landmarks that can be matched to large-scale geo-tagged image databases. Fig.~\ref{fig:twopic} illustrates a representative threat, where a released video frame can be linked to a geographically tagged reference image and thereby reveal the recording location.

This paper focuses on \emph{location privacy} under a background-based retrieval attacker. The setting is different from face or license-plate anonymization: even when explicit identities are hidden, non-human background cues such as buildings, road layouts, signage, and skylines can still expose where the video was recorded. We therefore study a more specific release question: how can a system publish or process useful video frames while reducing the attacker's ability to retrieve the location from background appearance?

\paragraph{Threat model and released object.}
Let $q$ denote a released video frame and $\mathcal{D}=\{(s_i,\ell_i)\}_{i=1}^N$ a gallery of geo-tagged reference images with location labels $\ell_i$. The attacker computes an embedding $f(\cdot)$ and ranks gallery images by similarity $\mathrm{sim}(f(q),f(s_i))$. A retrieval attack succeeds if the correct location appears in the Top-$k$ list. Our defense modifies the released frame before the attacker sees it. The downstream utility target is a detector or segmenter that still operates on the sanitized frame. In the controlled benchmark below, each query is a short sequence used to estimate a temporally smoothed mask, but retrieval is evaluated on the sanitized middle frame of that sequence rather than on a sequence-level aggregated descriptor.

\paragraph{Why not encryption?}
Encryption protects data in transit or at rest, but not the setting where decrypted frames must be processed for auditing, training, or analytics. Once the frame is available to a perception module, the same background cues are also available to a retrieval attacker. Our objective is therefore \emph{utility-preserving sanitization}: selectively perturb the background evidence that drives location retrieval while avoiding the heavy utility loss caused by global masking or indiscriminate noise.

\paragraph{Scope of the privacy claim.}
PPEDCRF uses a Gaussian-mechanism-inspired calibration to map a user-chosen privacy budget to a base noise scale, but we do \emph{not} claim a formal output-level $(\varepsilon,\delta)$-DP guarantee for the released video sequence. Instead, our contribution is a calibrated privacy--utility trade-off framework: PPEDCRF allows practitioners to continuously adjust the noise budget and select a desired operating point on the privacy--utility frontier. We evaluate this trade-off explicitly through matched-operating-point comparisons, temporal-consistency metrics, and attacker-sensitivity analyses.

\subsection{Related Work}
\textbf{Visual anonymization for multimedia.}
Visual privacy has been surveyed extensively~\cite{padilla2015visual,ribaric2016deidentification}, with classical methods focusing on obfuscating or removing foreground entities---primarily faces and license plates---through blurring, pixelation, or cryptographic protocols~\cite{Chamikara2020101951,Li2019212,avidan2007efficient,erkin2009privacy}. More recent generative approaches synthesize realistic de-identified faces: DeepPrivacy~\cite{hukkelaas2019deepprivacy} replaces faces via a GAN conditioned on sparse pose landmarks, while CIAGAN~\cite{maximov2020ciagan} performs conditional identity swaps that preserve expression and pose. Sun et al.~\cite{sun2018natural} propose head inpainting as a natural obfuscation mechanism that avoids the uncanny artifacts of na\"{\i}ve blurring. Oh et al.~\cite{oh2016faceless} demonstrated that modern recognition networks can re-identify individuals even when faces are fully occluded, underscoring the insufficiency of face-only anonymization. Zhou and Pun~\cite{zhou2020personal} extend pixelation to irrelevant bystanders in live video streams using tracking-based selection. Despite their effectiveness on foreground identities, none of these methods address the case where the sensitive signal lies in the \emph{background scene} that supports geo-localization---the threat model targeted by PPEDCRF.

\textbf{Privacy-preserving perception pipelines.}
A parallel line of research studies privacy-aware detection and recognition pipelines that modify model inputs, training procedures, or intermediate representations to limit information leakage. Liu et al.~\cite{liu2019privacy} adapt Faster R-CNN for privacy-preserving medical object detection, and Ren et al.~\cite{ren2018learning} learn face anonymization transformations that preserve action-detection accuracy. Kaissis et al.~\cite{kaissis2020secure} survey federated and encrypted inference for medical imaging, while Xu et al.~\cite{xu2019ganobfuscator} propose GANobfuscator, which applies GAN-based perturbations to images before sharing to prevent attribute inference. These works share a common emphasis on downstream task utility but adopt a foreground-centric threat model. Our setting differs because the primary privacy risk is the scene context that enables geo-localization rather than the identity of depicted persons.

\textbf{Visual place recognition and geo-localization.}
Visual place recognition (VPR) matches a query image to a database of geo-tagged reference images using learned global or local descriptors~\cite{lowry2016visual,masone2021survey}. NetVLAD~\cite{arandjelovic2016netvlad} introduced end-to-end trainable VLAD pooling over convolutional features extracted from deep backbones such as ResNet~\cite{he2016deep}, and 24/7 place recognition~\cite{torii201524} extended this to day--night and seasonal variations. Patch-NetVLAD~\cite{hausler2021patchnetvlad} further improves robustness through multi-scale patch-level descriptor fusion. Recent advances such as CosPlace~\cite{berton2022rethinking} and MixVPR~\cite{ali2023mixvpr} push retrieval accuracy higher through novel training losses and architectural innovations. These methods collectively define the attacker side of our threat model: the stronger and more diverse the VPR embedding, the harder it is for any defense to remain attacker-agnostic. Our updated evaluation includes three dedicated VPR embedders (CosPlace, MixVPR, Patch-NetVLAD) in addition to classification-style attackers, but it still does not cover the full design space (for example, NetVLAD variants and larger cross-view benchmarks). The current empirical evidence should therefore be interpreted as bounded transfer evidence rather than universal retrieval robustness.

\textbf{Scene-level and location privacy.}
Beyond foreground anonymization, several studies address scene-level information leakage in shared multimedia. Pittaluga et al.~\cite{pittaluga2019revealing} showed that point-cloud representations used for localization can leak detailed scene content, and Speciale et al.~\cite{speciale2019privacy} proposed privacy-preserving visual localization that hides geometric structure from the server. In the location-privacy literature, Andr\'{e}s et al.~\cite{andres2013geo} formalized geo-indistinguishability as a differential-privacy extension for location-based services, and Shokri et al.~\cite{shokri2011quantifying} proposed quantitative metrics for evaluating location privacy against inference attacks. These approaches are complementary to ours: they protect geometric or coordinate-level representations, whereas PPEDCRF protects the released pixel-domain video frame against image-retrieval-based geo-localization.

\textbf{Differential privacy and perturbation-based defenses.}
Differential privacy (DP) provides a rigorous mathematical framework for bounding information leakage~\cite{dwork2014algorithmic}. The Gaussian mechanism and its analytical calibration~\cite{balle2018improving} are widely used in practice, and R\'{e}nyi DP~\cite{mironov2017renyi} offers tighter composition bounds for iterative algorithms. Abadi et al.~\cite{abadi2016deep} introduced DP-SGD for training deep networks with formal privacy guarantees. In the multimedia domain, VideoDP~\cite{wang2020videodp} applies DP to video analytics, and Roy and Boddeti~\cite{roy2019mitigating} propose a maximum-entropy approach to mitigate information leakage in image representations. From the adversarial robustness perspective, perturbation-based defenses have been studied extensively: Goodfellow et al.~\cite{goodfellow2015explaining} introduced FGSM as both an attack and a defense mechanism, while Madry et al.~\cite{madry2018towards} proposed PGD-based adversarial training as a stronger defense. PPEDCRF adopts DP-style Gaussian calibration as a practical control knob for perturbation strength rather than claiming strict end-to-end DP for the released video; the connection to adversarial perturbation methods motivates future extensions toward adaptive, attacker-aware noise injection.

\textbf{Conditional random fields for spatial--temporal inference.}
CRFs have a long history in structured prediction for vision tasks. Kr\"{a}henb\"{u}hl and Koltun~\cite{krahenbuhl2011efficient} proposed efficient mean-field inference in fully connected CRFs with Gaussian edge potentials, enabling dense pixel-level labeling. Zheng et al.~\cite{zheng2015conditional} formulated CRF inference as a recurrent neural network layer, allowing end-to-end training of CRF-augmented segmentation models. Chen et al.~\cite{chen2017deeplab} integrated fully connected CRFs into the DeepLab semantic segmentation architecture, achieving state-of-the-art results on standard benchmarks. Dynamic CRFs~\cite{wang2005dynamic} extend the graphical model to video by incorporating temporal edges across frames, enabling consistent foreground--background labeling over time. PPEDCRF builds on this dynamic CRF tradition: it uses mean-field-style spatial smoothing combined with temporal consistency edges to produce stable, time-coherent sensitivity masks that guide selective perturbation.

\textbf{Privacy--utility trade-off optimization.}
A growing body of work explicitly studies the tension between privacy protection and downstream utility in image and video processing~\cite{wu2018towards}. McPherson et al.~\cite{mcpherson2016defeating} demonstrated that standard obfuscation techniques such as mosaicing and blurring can be defeated by deep learning models trained to invert the transformation, motivating the need for stronger and more principled perturbation strategies. Adversarial training frameworks learn privacy-protective transformations while preserving task accuracy, and optimization-based approaches balance obfuscation strength with perceptual quality. PPEDCRF contributes to this line by offering principled per-pixel calibration through DCRF-guided spatial support and NCP-controlled perturbation amplitude, together with an explicit frontier-based experimental evaluation that maps the achievable privacy--utility operating points.

\subsection{Contributions}
This paper positions PPEDCRF as a \emph{calibrated privacy--utility trade-off optimizer} for background-based location privacy in released video frames. Our contributions are:
\begin{itemize}
    \item A unified selective perturbation pipeline in which DCRF support localization identifies \emph{where} to perturb, NCP controls \emph{how strongly} to perturb within that support, and calibrated Gaussian noise implements the release-side mechanism in image space.
    \item A DP-style calibration rule that maps a user-chosen privacy budget to a continuously adjustable noise scale, explicitly distinguished from strict end-to-end differential privacy and interpreted as a practical operating-point control.
    \item A reproducible controlled paired-scene retrieval benchmark with matched-operating-point, temporal-consistency, and attacker-sensitivity analyses over both classification-style and dedicated VPR embedders, showing bounded attacker-dependent transfer and identifying support localization as the most consistently beneficial component.
\end{itemize}

\section{Method}
\label{sec:method}

\subsection{Pipeline Overview}
PPEDCRF operates on a released video sequence $\mathcal{X}=\{I_t\}_{t=1}^T$. For each frame, the method estimates a location-sensitive background map, smooths it temporally, normalizes the perturbation strength, and injects Gaussian noise only inside the inferred background regions. The implementation-consistent pipeline is
\[
I_t \rightarrow u_t \rightarrow p_t \rightarrow \alpha_t \rightarrow I_t',
\]
where $u_t$ is the unary sensitivity logit, $p_t$ is the DCRF-refined continuous mask, $\alpha_t$ is the NCP control map, and $I_t'$ is the released sanitized frame. The attacker then observes $I_t'$ and computes retrieval embeddings externally.

\begin{figure*}[!t]
    \centering
    \includegraphics[width=0.8\textwidth]{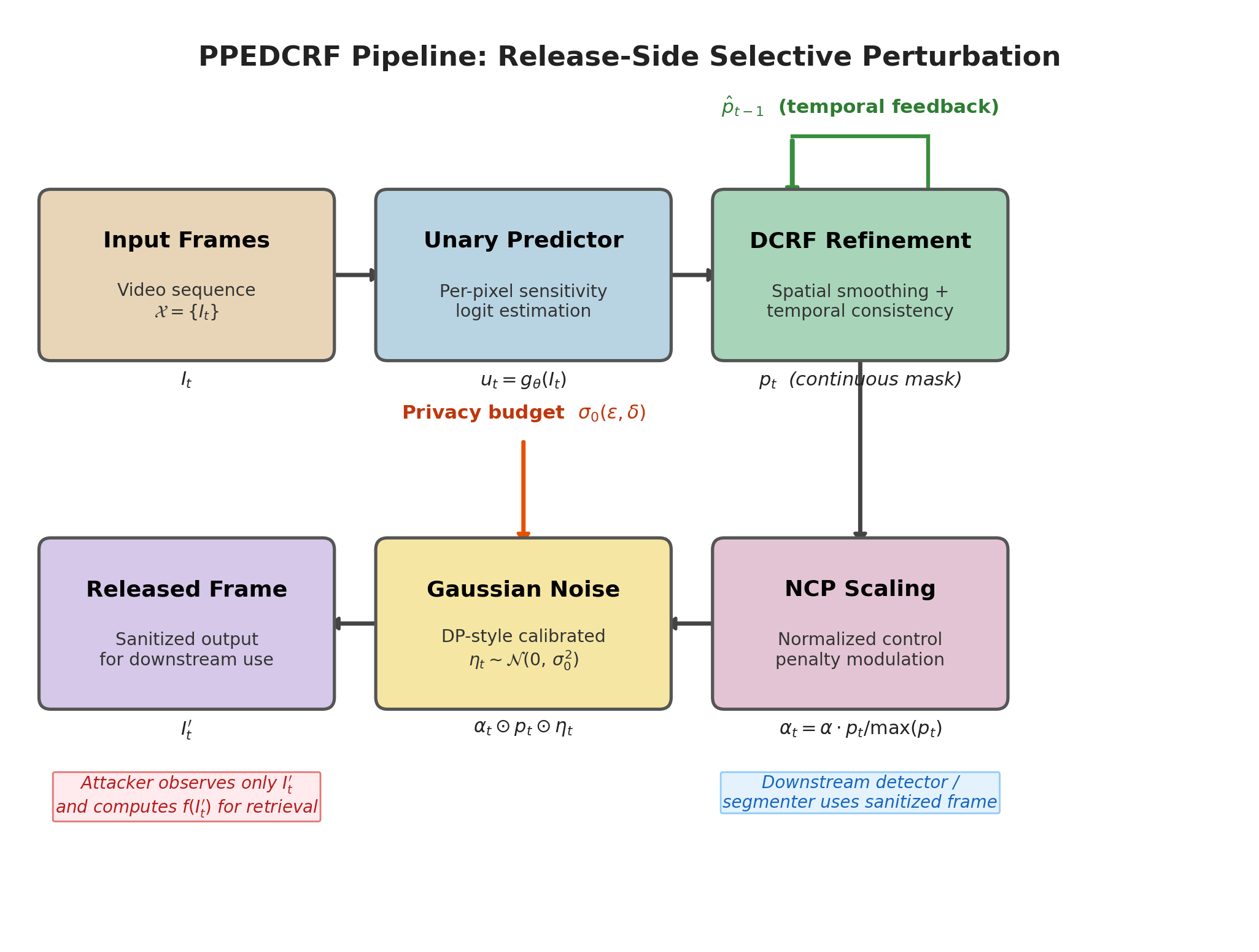}
    \caption{Overview of the PPEDCRF pipeline. A unary sensitive-region predictor produces per-frame logits, DCRF enforces spatial--temporal consistency, NCP rescales perturbation strength, and Gaussian noise is injected only in location-sensitive background regions before release.}
    \Description{A pipeline diagram with four stages: unary background-sensitivity prediction, dynamic CRF refinement across neighboring frames, normalized control penalty scaling, and selective Gaussian perturbation of the released frame.}
    \label{fig:arc}
\end{figure*}

\begin{table}[t]
\centering
\caption{Compact notation summary.}
\label{tab:symbols}
\begin{tabularx}{\linewidth}{>{\raggedright\arraybackslash}p{0.16\linewidth} X}
\hline
Symbol & Meaning \\
\hline
$I_t$ & Input frame at time $t$. \\
$u_t$ & Unary sensitivity logit predicted from $I_t$. \\
$p_t$ & DCRF-refined continuous sensitivity mask. \\
$\alpha_t$ & NCP control map that rescales perturbation strength. \\
$\sigma_0$ & Base Gaussian noise scale derived from the privacy budget. \\
$I_t'$ & Sanitized frame released to the attacker or downstream model. \\
$f(\cdot)$ & External retrieval backbone used by the attacker. \\
$\mathcal{D}$ & Retrieval gallery of candidate reference images. \\
\hline
\end{tabularx}
\end{table}

\subsection{DCRF Mask Inference}
The unary network $g_\theta(\cdot)$ predicts a per-pixel location-sensitivity logit
\begin{equation}
\label{eq:unary}
u_t = g_\theta(I_t).
\end{equation}
We initialize the continuous sensitivity map as $p_t^{(0)}=\sigma(u_t)$ and refine it with a dynamic CRF using a small number of mean-field iterations that matches the released implementation:
\begin{equation}
\label{eq:dcrf}
p_t^{(k+1)} =
\sigma\!\left(
u_t + \lambda_s \big(S(p_t^{(k)}) - p_t^{(k)}\big)
      + \lambda_\tau \big(\hat{p}_{t-1} - p_t^{(k)}\big)
\right),
\end{equation}
where $S(\cdot)$ is average-pooling-based spatial smoothing, $\hat{p}_{t-1}$ is the previous refined map warped into frame $t$ (or reused directly when optical flow is unavailable), and $\lambda_s,\lambda_\tau$ are spatial and temporal weights. After a fixed number of mean-field-style iterations, the final refined mask is denoted by $p_t$.

This formulation resolves a key ambiguity in older drafts: DCRF is not a separate theorem-heavy module detached from implementation. It is the mechanism that turns per-frame logits into a time-consistent background mask used immediately by the sanitizer.

\subsection{NCP Control and DP-Style Calibration}
The normalized control penalty (NCP) rescales the refined mask rather than redefining the threat model. In the released code, NCP is implemented as a normalized control map
\begin{equation}
\label{eq:ncp}
\alpha_t = \alpha \cdot \frac{p_t}{\max(p_t)+\epsilon},
\end{equation}
where $\alpha$ is a global control knob and $\epsilon$ avoids division by zero. Optional class-specific weights can be multiplied into $\alpha_t$ when semantic region labels are available; in the current experiments the control map is derived directly from the refined sensitivity mask. In other words, $p_t$ remains the spatial support and $\alpha_t$ only modulates the perturbation amplitude within that support.

We use Gaussian-mechanism-inspired calibration to map a user-selected privacy budget to the base noise level:
\begin{equation}
\label{eq:gaussian_calibration}
\sigma_0(\varepsilon,\delta) =
\frac{C\sqrt{2\log(1.25/\delta)}}{\varepsilon},
\end{equation}
where $C$ is an $\ell_2$ clipping bound on the per-pixel sensitivity. Equation~\eqref{eq:gaussian_calibration} is used only as a practical calibration rule for the perturbation strength. It is \emph{not} sufficient by itself to claim strict end-to-end differential privacy for the released video---a formal DP proof would additionally require bounding the composition over frames and verifying sensitivity with respect to individual training examples, which is outside the scope of this work.

\textbf{Why this calibration over simpler alternatives.}
The most natural alternatives are linear scaling ($\sigma_0 \propto 1/\varepsilon$) and a fixed constant schedule. Linear scaling grows unboundedly as $\varepsilon \to 0$ without providing any probabilistic control on the $\delta$ parameter, while a fixed schedule has no principled connection to the user's desired privacy level. The Gaussian-mechanism form in Eq.~\eqref{eq:gaussian_calibration} has three practical properties that motivate its use: (i) it is monotonically decreasing in $\varepsilon$ and in $\delta$, so smaller budgets always map to larger noise as expected; (ii) the $\sqrt{2\log(1.25/\delta)}$ factor provides a concrete interpretation of $\delta$---the probability tolerance on exceeding the nominal protection level---making it easier for practitioners to select sensible parameters; (iii) it degrades continuously to linear scaling as $\delta \to 1/1.25$, providing a smooth bridge to the simpler rule. In our experiments, we fix $\delta=10^{-5}$ so that $\sqrt{2\log(1.25/\delta)}\approx 4.73$, and the clipping bound $C$ is set equal to the normalized pixel range so that $\sigma_0$ has a direct interpretation in image-space units. This means $\sigma_0=8$ corresponds to $(\varepsilon, \delta)\approx(0.59, 10^{-5})$ under Eq.~\eqref{eq:gaussian_calibration}, serving as a concrete anchor point for the frontier in Fig.~\ref{fig:tradeoff}.

\subsection{Selective Image-Space Perturbation}
After DCRF and NCP, PPEDCRF perturbs only the inferred background regions:
\begin{equation}
\label{eq:image_noise}
I_t' = \mathrm{clip}\!\left(I_t + \alpha_t \odot p_t \odot \eta_t,\ 0,\ 255\right),
\qquad
\eta_t \sim \mathcal{N}(0,\sigma_0^2 I).
\end{equation}
Equation~\eqref{eq:image_noise} is the released mechanism evaluated in this paper. The attacker never observes intermediate maps such as $u_t$, $p_t$, or $\alpha_t$; it only observes the sanitized frame $I_t'$ and computes its own embedding $f(I_t')$ for retrieval. The product $\alpha_t \odot p_t$ is intentional: $p_t$ defines where perturbation is allowed, while $\alpha_t$ further increases the amplitude on the highest-confidence sensitive pixels inside that same support.

Selective perturbation matters because global Gaussian noise damages foreground perception disproportionately, while random masking perturbs many pixels that are irrelevant to location retrieval. PPEDCRF instead concentrates perturbation energy on background regions that remain stable across adjacent frames and are therefore more likely to support geo-localization.

\subsection{Algorithm}
\begin{algorithm}[t]
\caption{PPEDCRF Release-Side Sanitization}
\label{alg:ppedcrf}
\begin{algorithmic}[1]
\Statex \textbf{Input:} sequence $\mathcal{X}=\{I_t\}_{t=1}^T$, unary predictor $g_\theta$, DCRF parameters $(\lambda_s,\lambda_\tau)$, NCP scale $\alpha$, base noise scale $\sigma_0$
\Statex \textbf{Output:} sanitized sequence $\mathcal{X}'=\{I_t'\}_{t=1}^T$
\State $\hat{p}_0 \gets \mathbf{0}$
\For{$t=1$ to $T$}
    \State $u_t \gets g_\theta(I_t)$
    \State $p_t \gets \mathrm{DCRFRefine}(u_t,\hat{p}_{t-1})$
    \State $\alpha_t \gets \alpha \cdot p_t / (\max(p_t)+\epsilon)$
    \State Sample $\eta_t \sim \mathcal{N}(0,\sigma_0^2 I)$
    \State $I_t' \gets \mathrm{clip}(I_t + \alpha_t \odot p_t \odot \eta_t, 0, 255)$
    \State $\hat{p}_t \gets p_t$
\EndFor
\State \Return $\mathcal{X}'$
\end{algorithmic}
\end{algorithm}

On a single monitoring frame at $192\times 320$ resolution, DCRF mask inference and noise injection complete in under 50\,ms on a consumer GPU (NVIDIA GTX 1060), making the pipeline compatible with near-real-time release workflows.

\section{Experimental Evaluation}
\label{sec:experiments}

\subsection{Experimental Protocol}
We report two complementary experiment tracks.

\textbf{Legacy driving-dataset utility results.}
Legacy detector-training outputs on MOT16/17, Cityscapes, KITTI, and VOC2008 are retained as auxiliary downstream-utility evidence and moved to Appendix for space efficiency. They show that sanitization does not catastrophically collapse legacy perception workloads, but they are not the primary evidence for the location-retrieval threat model.

\textbf{New controlled paired-scene retrieval benchmark.}
To directly evaluate the background-based retrieval attacker under a fully reproducible setting, we add a controlled benchmark built from local monitoring sequences. We first index 120 candidate sequences and extract one representative frame from each. Using a fixed ResNet18~\cite{he2016deep} embedder, we greedily pair the most similar sequences into 12 synthetic location pairs; one sequence supplies gallery views and the paired sequence supplies query views. We then add 36 hard distractors selected by maximum similarity to the paired locations, yielding galleries of size 12, 24, and 48. All results are averaged over three noise seeds, and the benchmark is generated by the released script \texttt{src/scripts/run\_controlled\_retrieval\_benchmark.py}.

The mined benchmark is intentionally difficult rather than random: across the 12 retained gallery--query pairs, the anchor ResNet18 cosine similarity averages $\approx$0.99 (range 0.98--1.00), and the 36 hard distractors average 0.86 maximum similarity to the paired set (hardest: 0.99). We therefore treat the benchmark as a hard-negative proxy for location retrieval rather than as a gallery of arbitrary negatives.

The two experiment tracks play different roles in the paper. The legacy detector and segmenter outputs reflect the original training-time utility pipeline, whereas the controlled retrieval benchmark is an inference-time evaluation with fixed attacker backbones and sanitized query views.

\begin{table}[t]
\centering
\caption{Datasets and roles in the revised evaluation.}
\label{tab:datasets}
\begin{tabular}{p{0.24\linewidth} p{0.68\linewidth}}
\hline
Dataset / benchmark & Role in this paper \\
\hline
MOT16 / MOT17, Cityscapes, KITTI, VOC2008 & Legacy utility references moved to Appendix. \\
Controlled paired-scene benchmark & New retrieval, ablation, privacy--utility frontier, and attacker-sensitivity evaluation built from monitoring sequences. \\
\hline
\end{tabular}
\end{table}

Unless otherwise stated, frames are resized to $192\times 320$, the retrieval attacker uses cosine similarity on fixed image embeddings, and the reported privacy metrics are Top-1, Top-5, and Top-10 retrieval accuracy. Each query sequence contributes one sanitized middle frame for attack evaluation, while the surrounding frames are used only to stabilize the DCRF mask estimate. Perceptual utility is measured with PSNR and SSIM~\cite{wang2004image} between original and sanitized query frames. Raw queries are deterministic under the fixed gallery. All sanitized variants are averaged over three independent noise seeds (1234, 1235, 1236). To improve comparability across attacker families, all backbone-specific results are computed from one unified benchmark construction and the same three-seed averaging protocol---a deliberate methodological choice to eliminate earlier ambiguity in cross-backbone comparison.

\textbf{Scope of this benchmark.} The paired-scene benchmark is designed to isolate release-side privacy mechanisms when a full geo-localization dataset is not locally available. It is anchored on ResNet18 similarity when mining query--gallery pairs, which improves reproducibility but may bias the benchmark toward retrieval models whose features resemble the anchor embedder. The added ResNet50/VGG16 transfer and dedicated VPR transfer (CosPlace, MixVPR, Patch-NetVLAD) partly test this issue, but they do not fully eliminate benchmark-construction bias. Results support claims about privacy--utility trade-offs, masking strategy, and attacker sensitivity, but they do not by themselves prove universal robustness against every retrieval backbone or constitute a formal differential-privacy certificate. We discuss this distinction further in Appendix~A, which also provides implementation-alignment notes and a larger-scale 50-pair confirmation used to assess whether the main 12-pair conclusions persist under a broader evaluation set.

\textbf{Comparability across reported tables.} All tables now share the same unified benchmark run with three noise seeds and all eight attacker backbones, produced from a single output directory (\texttt{controlled\_retrieval\_seed\_avg}). Table~\ref{tab:ablation}, Table~\ref{tab:sigma_sweep}, and Table~\ref{tab:matched} focus on mechanism behavior under the default backbone (ResNet18), while Table~\ref{tab:robustness} reports cross-backbone transfer. Because the same pair discovery, gallery construction, and seed set are shared across all tables, cross-table comparisons are directly meaningful.

\subsection{Legacy Utility Track (Appendix Summary)}
Legacy detector and segmentation experiments (MOT16/17, Cityscapes, KITTI, VOC2008) are reported in the Appendix. They confirm that selective perturbation preserves downstream perception behavior but do not directly measure location-retrieval privacy. The main conclusions below rely on the paired-scene benchmark.

\subsection{Controlled Paired-Scene Retrieval Benchmark}
Table~\ref{tab:ablation} reports the default ablation setting: ResNet18 attacker, gallery size 48, and $\sigma_0=8$. The raw query baseline reaches Top-1 retrieval accuracy 0.667. PPEDCRF lowers this to $0.361\pm0.127$ while preserving $36.14\pm0.01$\,dB PSNR and $0.918\pm0.001$ SSIM. Global Gaussian noise achieves the strongest raw privacy ($0.083$ Top-1) but at substantially lower visual quality (30.17\,dB PSNR). A random mask achieves Top-1 $0.278\pm0.048$ with lower perceptual quality (33.15\,dB PSNR) and much poorer temporal consistency (mask IoU 0.334). Among the support-aware baselines, mask-guided blur reaches Top-1 0.500 at 30.47\,dB PSNR---now worse than PPEDCRF on both privacy and quality---while mask-guided mosaic reaches Top-1 0.167 at 25.03\,dB PSNR with a large quality cost. On this benchmark, PPEDCRF provides the strongest high-utility privacy--utility operating point among the stochastic methods evaluated here, and it also improves over mask-guided blur in absolute privacy, while deterministic support-aware baselines occupy complementary lower-utility operating regions.

Top-10 retrieval saturates at 1.000 for all variants in this 48-candidate proxy setting, so Table~\ref{tab:ablation} reports the more discriminative Top-1 and Top-5 values in the main paper while keeping the full Top-10 summaries in the released CSV files.

The DCRF and NCP ablations are informative for a different reason. On these short monitoring sequences, removing temporal consistency or replacing NCP with a fixed control map changes Top-1 only marginally, and temporal metrics remain nearly identical across PPEDCRF, w/o temporal, and w/o NCP (Section~\ref{sec:temporal}). Together with the strong blur and mosaic baselines, this suggests that the local paired-scene benchmark mainly stresses \emph{where} the perturbation is applied; the temporal and calibration modules remain methodologically meaningful but are only weakly distinguished by the current short-clip protocol.

\begin{table*}[t]
\centering
\caption{Controlled paired-scene benchmark, default setting: ResNet18 attacker, gallery size 48, $\sigma_0=8$, 12 paired locations, and 36 hard distractors. Lower retrieval accuracy indicates stronger privacy. Each query metric is computed on the sanitized middle frame of a short query sequence. Stochastic variants report mean$\pm$std over three noise seeds; raw and deterministic baselines (blur, mosaic) are exact.}
\label{tab:ablation}
\begin{tabular}{lccccc}
\hline
Method & Top-1 $\downarrow$ & Top-5 $\downarrow$ & PSNR (dB) $\uparrow$ & SSIM $\uparrow$ & Mask IoU $\uparrow$ \\
\hline
Raw query & 0.667 & 1.000 & -- & -- & -- \\
PPEDCRF & $0.361\pm0.127$ & $0.917\pm0.000$ & $36.14\pm0.01$ & $0.918\pm0.001$ & 0.998 \\
w/o temporal consistency & $0.361\pm0.127$ & $0.917\pm0.000$ & $36.14\pm0.01$ & $0.918\pm0.001$ & 1.000 \\
w/o NCP (fixed sigma) & $0.361\pm0.127$ & $0.917\pm0.000$ & $36.13\pm0.01$ & $0.918\pm0.001$ & 0.998 \\
Mask-guided blur & $0.500$ & $0.917$ & $30.47$ & $0.941$ & 0.998 \\
Mask-guided mosaic & $0.167$ & $0.500$ & $25.03$ & $0.899$ & 0.998 \\
Random mask & $0.278\pm0.048$ & $0.750\pm0.083$ & $33.15\pm0.01$ & $0.857\pm0.001$ & 0.334 \\
Global Gaussian noise & $0.083\pm0.000$ & $0.611\pm0.096$ & $30.17\pm0.01$ & $0.754\pm0.001$ & 1.000 \\
\hline
\end{tabular}
\end{table*}

Table~\ref{tab:sigma_sweep} extends the ablation across increasing noise budgets $\sigma_0\in\{8,16,24,32\}$. The three PPEDCRF variants (full, w/o temporal, w/o NCP) remain indistinguishable in Top-1 at every $\sigma_0$, confirming that the privacy effect is driven by the spatial support rather than the temporal or NCP modules. As $\sigma_0$ increases from 8 to 32, Top-1 decreases from $0.361$ to $0.083$ while PSNR drops from 36.14\,dB to 24.22\,dB---a trajectory that matches the frontier in Fig.~\ref{fig:tradeoff}.

\begin{table}[t]
\centering
\caption{Ablation across noise budgets. All variants share the same DCRF support; differences appear only in quality metrics. ResNet18 attacker, gallery 48. All entries are stochastic (three seeds) and report mean$\pm$std.}
\label{tab:sigma_sweep}
\begin{tabular}{lccccc}
\hline
$\sigma_0$ & Variant & Top-1 $\downarrow$ & Top-5 $\downarrow$ & PSNR $\uparrow$ & SSIM $\uparrow$ \\
\hline
8 & PPEDCRF & $0.361\pm0.127$ & $0.917\pm0.000$ & $36.14\pm0.01$ & $0.918\pm0.001$ \\
8 & w/o temporal & $0.361\pm0.127$ & $0.917\pm0.000$ & $36.14\pm0.01$ & $0.918\pm0.001$ \\
8 & w/o NCP & $0.361\pm0.127$ & $0.917\pm0.000$ & $36.13\pm0.01$ & $0.918\pm0.001$ \\
\hline
16 & PPEDCRF & $0.083\pm0.000$ & $0.611\pm0.096$ & $30.16\pm0.01$ & $0.754\pm0.001$ \\
16 & w/o temporal & $0.083\pm0.000$ & $0.611\pm0.096$ & $30.15\pm0.01$ & $0.754\pm0.001$ \\
16 & w/o NCP & $0.083\pm0.000$ & $0.611\pm0.096$ & $30.15\pm0.01$ & $0.754\pm0.001$ \\
\hline
24 & PPEDCRF & $0.083\pm0.000$ & $0.444\pm0.048$ & $26.67\pm0.01$ & $0.597\pm0.001$ \\
24 & w/o temporal & $0.083\pm0.000$ & $0.444\pm0.048$ & $26.67\pm0.01$ & $0.597\pm0.001$ \\
24 & w/o NCP & $0.083\pm0.000$ & $0.444\pm0.048$ & $26.67\pm0.01$ & $0.597\pm0.001$ \\
\hline
32 & PPEDCRF & $0.083\pm0.000$ & $0.389\pm0.048$ & $24.22\pm0.01$ & $0.477\pm0.001$ \\
32 & w/o temporal & $0.083\pm0.000$ & $0.417\pm0.083$ & $24.22\pm0.01$ & $0.477\pm0.001$ \\
32 & w/o NCP & $0.083\pm0.000$ & $0.389\pm0.048$ & $24.21\pm0.01$ & $0.476\pm0.001$ \\
\hline
\end{tabular}
\end{table}

\subsection{Privacy--Utility Frontier, Attacker Sensitivity, and Failure Modes}
\begin{figure*}[!t]
    \centering
    \includegraphics[width=0.8\textwidth]{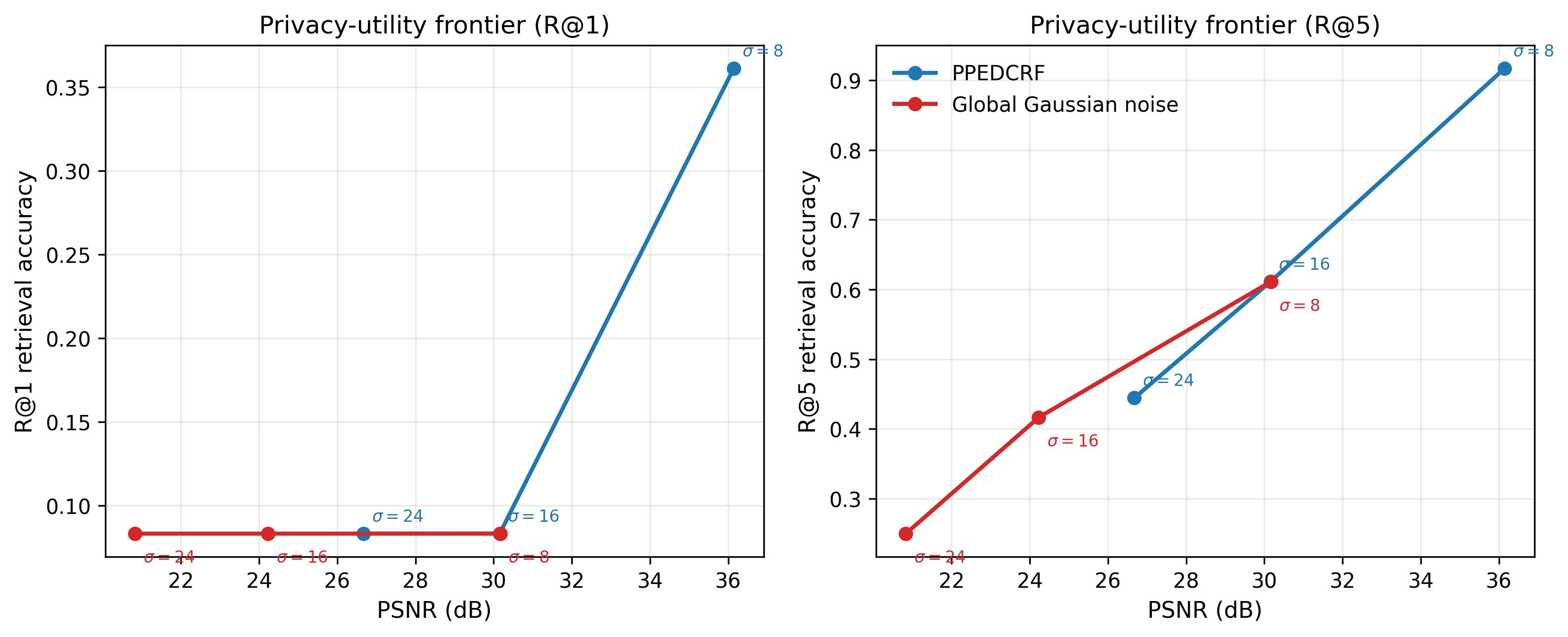}
    \caption{Privacy--utility frontier. Lower retrieval accuracy means better privacy; higher PSNR means better visual utility. PPEDCRF preserves an approximately 6\,dB PSNR advantage over indiscriminate global Gaussian noise across all tested privacy levels.}
    \Description{A privacy-utility trade-off plot showing retrieval accuracy versus PSNR for PPEDCRF and global Gaussian noise at increasing noise scales. PPEDCRF retains higher PSNR at comparable privacy levels.}
    \label{fig:tradeoff}
\end{figure*}

Figure~\ref{fig:tradeoff} sweeps the noise scale $\sigma_0\in\{8,16,24\}$ on the same paired-scene benchmark. The figure confirms PPEDCRF's core practical advantage: at every noise level, selective perturbation preserves substantially more perceptual quality than global Gaussian noise. At $\sigma_0=16$, both PPEDCRF and global Gaussian noise converge to Top-1 $0.083$, but PPEDCRF preserves $30.16$\,dB PSNR versus $24.23$\,dB for global noise---a gap of approximately 6\,dB at the same achieved privacy. At $\sigma_0=24$, the same pattern holds: both methods maintain Top-1 $0.083$ while PPEDCRF yields $26.67$\,dB versus $20.83$\,dB (all global-noise PSNR values are from the frontier data plotted in Fig.~\ref{fig:tradeoff} and available in the released benchmark outputs). This consistent $\approx$6\,dB PSNR advantage across noise budgets demonstrates that PPEDCRF delivers a materially better privacy--utility trade-off curve by concentrating perturbation energy on the DCRF-inferred support rather than spreading it uniformly.

\begin{figure*}[!t]
    \centering
    \begin{subfigure}[t]{0.95\textwidth}
        \centering
        \includegraphics[width=\linewidth]{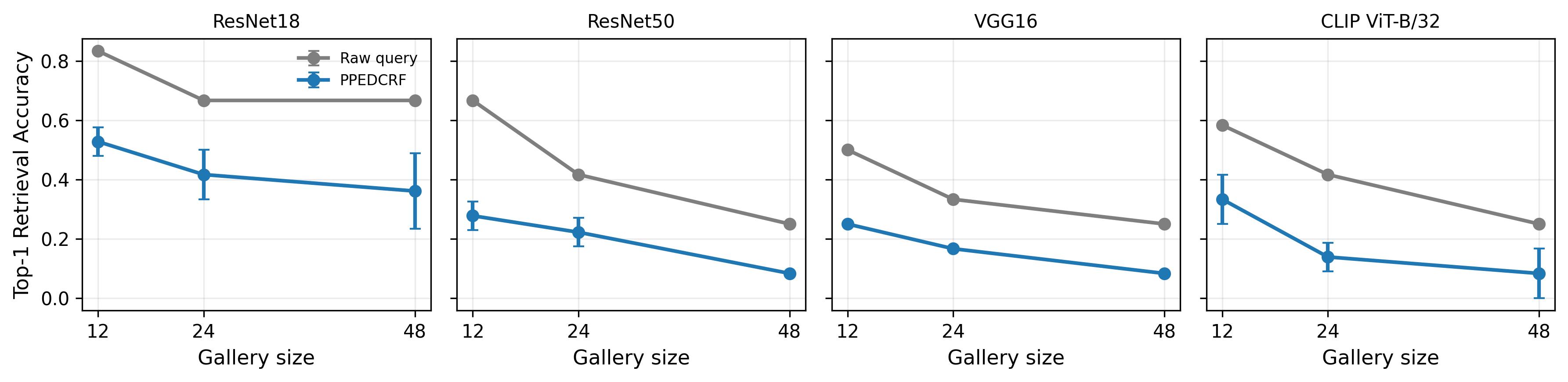}
        \caption{Top row: ResNet18, ResNet50, VGG16, and CLIP ViT-B/32.}
    \end{subfigure}

    \vspace{0.4em}

    \begin{subfigure}[t]{0.95\textwidth}
        \centering
        \includegraphics[width=\linewidth]{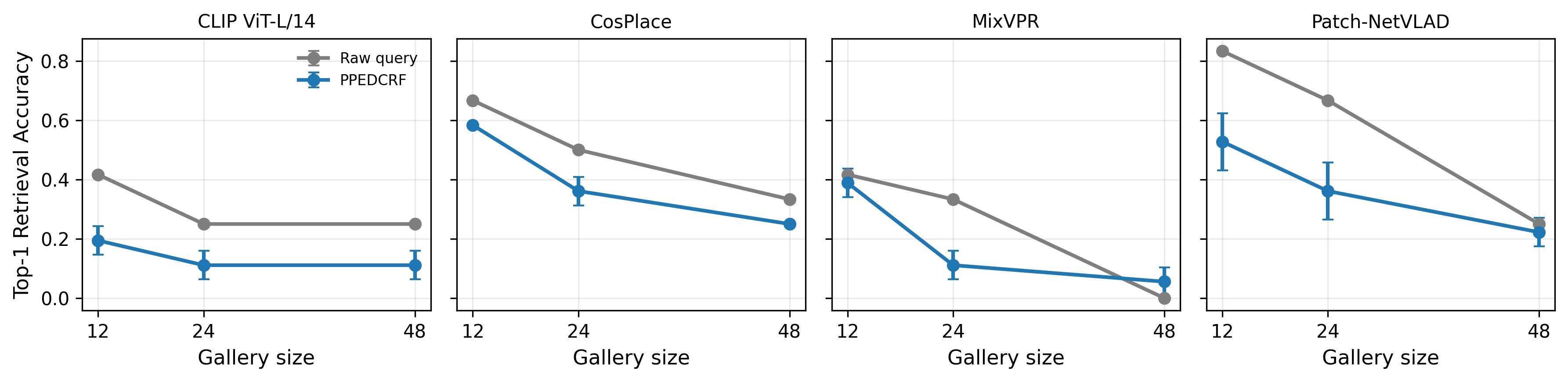}
        \caption{Bottom row: CLIP ViT-L/14, CosPlace, MixVPR, and Patch-NetVLAD.}
    \end{subfigure}

    \caption{Attacker-backbone and gallery-size sensitivity from the unified seed-averaged eight-backbone benchmark (three seeds), rendered as two stacked four-panel subfigures for readability. Across all eight backbones, PPEDCRF reduces Top-1 in 23 of 24 cells; the sole exception is MixVPR at gallery 48 where the raw baseline is already zero. CLIP ViT-L/14, which appeared adverse under a single-run protocol, is now consistently supportive under seed averaging.}
    \Description{Two stacked four-panel line charts. The top subfigure shows ResNet18, ResNet50, VGG16, and CLIP ViT-B/32. The bottom subfigure shows CLIP ViT-L/14, CosPlace, MixVPR, and Patch-NetVLAD. Each panel compares raw and PPEDCRF Top-1 retrieval accuracy across gallery sizes 12, 24, and 48, with seed-averaged confidence intervals.}
    \label{fig:robustness}
\end{figure*}

Table~\ref{tab:robustness} and Fig.~\ref{fig:robustness} vary gallery size and attacker backbone using the unified seed-averaged benchmark (three seeds, eight backbones, one output directory). Fig.~\ref{fig:robustness} visualizes the same gallery sizes as Table~\ref{tab:robustness} (12, 24, and 48 candidates from the main benchmark); it does not depict the larger-scale appendix protocol. Under ResNet18, PPEDCRF lowers Top-1 from 0.833 to $0.528\pm0.048$ (gallery 12) and from 0.667 to $0.361\pm0.127$ (gallery 48). ResNet50 and VGG16 show consistent improvement, with ResNet50 reaching $\Delta=-0.389$ at gallery 12 and VGG16 reaching $\Delta=-0.250$ at gallery 12. CLIP behavior is now consistently supportive under seed averaging: CLIP ViT-B/32 shows $\Delta=-0.250$ at gallery 12 and $-0.167$ at gallery 48, while CLIP ViT-L/14---previously adverse in single-run results---now shows negative deltas across all galleries ($\Delta=-0.222$ at gallery 12, $-0.139$ at galleries 24/48). Dedicated VPR embedders remain supportive: Patch-NetVLAD shows the strongest improvement ($\Delta=-0.306$ at galleries 12 and 24), CosPlace is modest but consistent, and MixVPR exhibits one marginal positive delta ($+0.056$ at gallery 48, where the raw baseline is already $0.000$).

Across all 24 backbone--gallery cells in Table~\ref{tab:robustness}, 23 show negative $\Delta$ and only one is marginally positive (MixVPR at gallery 48). Compared to the prior single-run results where CLIP ViT-L/14 appeared adverse, seed averaging substantially reduces this instability: the earlier positive deltas were artifacts of a single noise realization rather than a systematic failure mode, though MixVPR remains a consistent adverse-transfer exception.

At the same time, attacker transfer remains heterogeneous in magnitude. CosPlace shows the smallest deltas (as low as $-0.083$), and MixVPR's gallery-48 cell is marginally positive from a raw baseline of zero. A scaling confirmation with 50 paired locations and galleries up to 100 (Appendix) corroborates this pattern: 21 of 24 larger-benchmark cells are negative, while MixVPR shows consistent adverse transfer across all three gallery sizes. We therefore interpret these gains as broadly supportive transfer across diverse attacker families under the current benchmark protocol, while noting that MixVPR remains a consistent adverse-transfer exception. While the main benchmark remains a controlled paired-scene proxy rather than a large GPS-grounded geolocalization dataset, the appendix-level 50-pair scaling confirmation preserves the same qualitative pattern, indicating that the main findings are not purely an artifact of the small 12-pair setting.

Taken together, the seed-averaged main benchmark and the appendix-scale confirmation support the same qualitative conclusion: PPEDCRF exhibits broadly supportive transfer under the current release-side evaluation protocol, but the transfer remains bounded and attacker-dependent.

\begin{table*}[t]
\centering
\caption{Retrieval robustness across gallery size and attacker backbone ($\sigma_0=8$, three seeds). Lower Top-1 indicates stronger privacy. $\Delta$ = PPEDCRF $-$ raw; negative values indicate privacy improvement. Raw and deterministic values are exact; stochastic variants report mean$\pm$std over three noise seeds from a unified benchmark run with all eight backbones (\texttt{controlled\_retrieval\_seed\_avg}).}
\label{tab:robustness}
\begin{tabular}{lccccc}
\hline
Backbone & Gallery & Raw Top-1 $\downarrow$ & PPEDCRF Top-1 $\downarrow$ & $\Delta$ \\
\hline
ResNet18 & 12 & 0.833 & $0.528\pm0.048$ & $-0.306$ \\
ResNet18 & 24 & 0.667 & $0.417\pm0.083$ & $-0.250$ \\
ResNet18 & 48 & 0.667 & $0.361\pm0.127$ & $-0.306$ \\
\hline
ResNet50 & 12 & 0.667 & $0.278\pm0.048$ & $-0.389$ \\
ResNet50 & 24 & 0.417 & $0.222\pm0.048$ & $-0.194$ \\
ResNet50 & 48 & 0.250 & $0.083\pm0.000$ & $-0.167$ \\
\hline
VGG16 & 12 & 0.500 & $0.250\pm0.000$ & $-0.250$ \\
VGG16 & 24 & 0.333 & $0.167\pm0.000$ & $-0.167$ \\
VGG16 & 48 & 0.250 & $0.083\pm0.000$ & $-0.167$ \\
\hline
CLIP ViT-B/32 & 12 & 0.583 & $0.333\pm0.083$ & $-0.250$ \\
CLIP ViT-B/32 & 24 & 0.417 & $0.139\pm0.048$ & $-0.278$ \\
CLIP ViT-B/32 & 48 & 0.250 & $0.083\pm0.083$ & $-0.167$ \\
\hline
CLIP ViT-L/14 & 12 & 0.417 & $0.194\pm0.048$ & $-0.222$ \\
CLIP ViT-L/14 & 24 & 0.250 & $0.111\pm0.048$ & $-0.139$ \\
CLIP ViT-L/14 & 48 & 0.250 & $0.111\pm0.048$ & $-0.139$ \\
\hline
CosPlace & 12 & 0.667 & $0.583\pm0.000$ & $-0.083$ \\
CosPlace & 24 & 0.500 & $0.361\pm0.048$ & $-0.139$ \\
CosPlace & 48 & 0.333 & $0.250\pm0.000$ & $-0.083$ \\
\hline
MixVPR & 12 & 0.417 & $0.389\pm0.048$ & $-0.028$ \\
MixVPR & 24 & 0.333 & $0.111\pm0.048$ & $-0.222$ \\
MixVPR & 48 & 0.000 & $0.056\pm0.048$ & $+0.056$ \\
\hline
Patch-NetVLAD & 12 & 0.833 & $0.528\pm0.096$ & $-0.306$ \\
Patch-NetVLAD & 24 & 0.667 & $0.361\pm0.096$ & $-0.306$ \\
Patch-NetVLAD & 48 & 0.250 & $0.222\pm0.048$ & $-0.028$ \\
\hline
\end{tabular}
\end{table*}

The robustness analysis exposes important patterns for interpretation. First, PPEDCRF now shows consistent transfer to dedicated VPR embedders under seed averaging: CosPlace, MixVPR, and Patch-NetVLAD all exhibit lower mean Top-1 after sanitization across most gallery sizes. Second, the CLIP ViT-L/14 adverse transfer previously observed under single-run evaluation is substantially reduced: seed averaging reveals that all three CLIP ViT-L/14 gallery cells are negative-$\Delta$, confirming that the earlier positive deltas were stochastic artifacts rather than a systematic failure. Third, the benchmark remains rank-sensitive rather than fully obfuscating: Top-10 is not fully suppressed under all settings, and stronger attacker-adaptive perturbations would still be needed for more decisive privacy guarantees.

\subsection{Matched-Operating-Point Analysis}
\label{sec:matched}

A fair comparison between perturbation strategies requires controlling for perceptual quality. Table~\ref{tab:matched} summarizes three matched-utility conditions drawn from the frontier in Fig.~\ref{fig:tradeoff}.

\textbf{High quality ($\approx$36\,dB PSNR).}
When the utility target is high, both PPEDCRF ($\sigma_0=8$) and global Gaussian noise ($\sigma_0=4$) achieve identical Top-1 ($0.361\pm0.127$) at the same PSNR. At this operating point, the perturbation is weak enough that feature rankings change minimally, so the two noise-based methods converge. Mask-guided blur at its default parameters reaches only 30.47\,dB PSNR with Top-1 0.500, so it falls short of this quality target and provides weaker privacy.

\textbf{Moderate quality ($\approx$33\,dB PSNR).}
PPEDCRF at $\sigma_0=12$ and global noise at $\sigma_0=6$ both achieve Top-1 $0.250\pm0.083$ at $\approx$32.65\,dB. Again the stochastic methods converge when calibrated to the same quality.

\textbf{Lower quality ($\approx$30\,dB PSNR).}
At this stronger perturbation level, PPEDCRF at $\sigma_0=16$ achieves Top-1 0.083 at 30.16\,dB---identical to global noise at $\sigma_0=8$ (Top-1 0.083, 30.17\,dB). The deterministic baselines provide an informative reference: blur achieves Top-1 0.500 at 30.47\,dB (substantially weaker privacy) and mosaic reaches 0.167 at 25.03\,dB, confirming that stochastic Gaussian noise disrupts retrieval features more effectively than deterministic blur at similar perceptual degradation levels.

The consistent pattern across all three panels is that when PPEDCRF and global Gaussian noise are calibrated to the same PSNR, they achieve the same Top-1 privacy. PPEDCRF's practical benefit is therefore not stronger matched-utility privacy but rather that at any given noise scale~$\sigma_0$, selective perturbation preserves approximately 6\,dB more PSNR than global perturbation by concentrating noise energy on location-sensitive regions instead of uniformly degrading all image content. The DCRF contribution is clearest in support localization; under the current short-clip protocol, the temporal component mainly defines a structured support process, but its quantitative effect is only weakly separated from the temporal ablation.

\begin{table}[t]
\centering
\caption{Matched-operating-point comparison (ResNet18, gallery 48). Each panel matches methods at a target PSNR. Blur and mosaic use fixed parameters on the same DCRF support (their PSNR is independent of $\sigma_0$). Stochastic variants report mean$\pm$std over three noise seeds; deterministic baselines are exact. PSNR and SSIM columns report stochastic means; their seed-to-seed standard deviations ($\leq$0.01\,dB and $\leq$0.001 respectively) are omitted for readability.}
\label{tab:matched}
\begin{tabular}{lcccc}
\hline
Method & $\sigma_0$ & Top-1 $\downarrow$ & PSNR (dB) $\uparrow$ & SSIM $\uparrow$ \\
\hline
\multicolumn{5}{l}{\textit{A.\;High quality ($\approx$36\,dB PSNR)}} \\
PPEDCRF & 8 & $0.361\pm0.127$ & 36.14 & 0.918 \\
Global Gaussian noise & 4 & $0.361\pm0.127$ & 36.15 & 0.918 \\
Mask-guided blur & -- & 0.500 & 30.47 & 0.941 \\
\hline
\multicolumn{5}{l}{\textit{B.\;Moderate quality ($\approx$33\,dB PSNR)}} \\
PPEDCRF & 12 & $0.250\pm0.083$ & 32.64 & 0.833 \\
Global Gaussian noise & 6 & $0.250\pm0.083$ & 32.65 & 0.833 \\
Mask-guided blur & -- & 0.500 & 30.47 & 0.941 \\
\hline
\multicolumn{5}{l}{\textit{C.\;Lower quality ($\approx$30\,dB PSNR)}} \\
PPEDCRF & 16 & $0.083\pm0.000$ & 30.16 & 0.754 \\
Global Gaussian noise & 8 & $0.083\pm0.000$ & 30.17 & 0.754 \\
Mask-guided blur & -- & 0.500 & 30.47 & 0.941 \\
Mask-guided mosaic & -- & 0.167 & 25.03 & 0.899 \\
\hline
\end{tabular}
\end{table}

\begin{figure}[t]
    \centering
    \includegraphics[width=\linewidth]{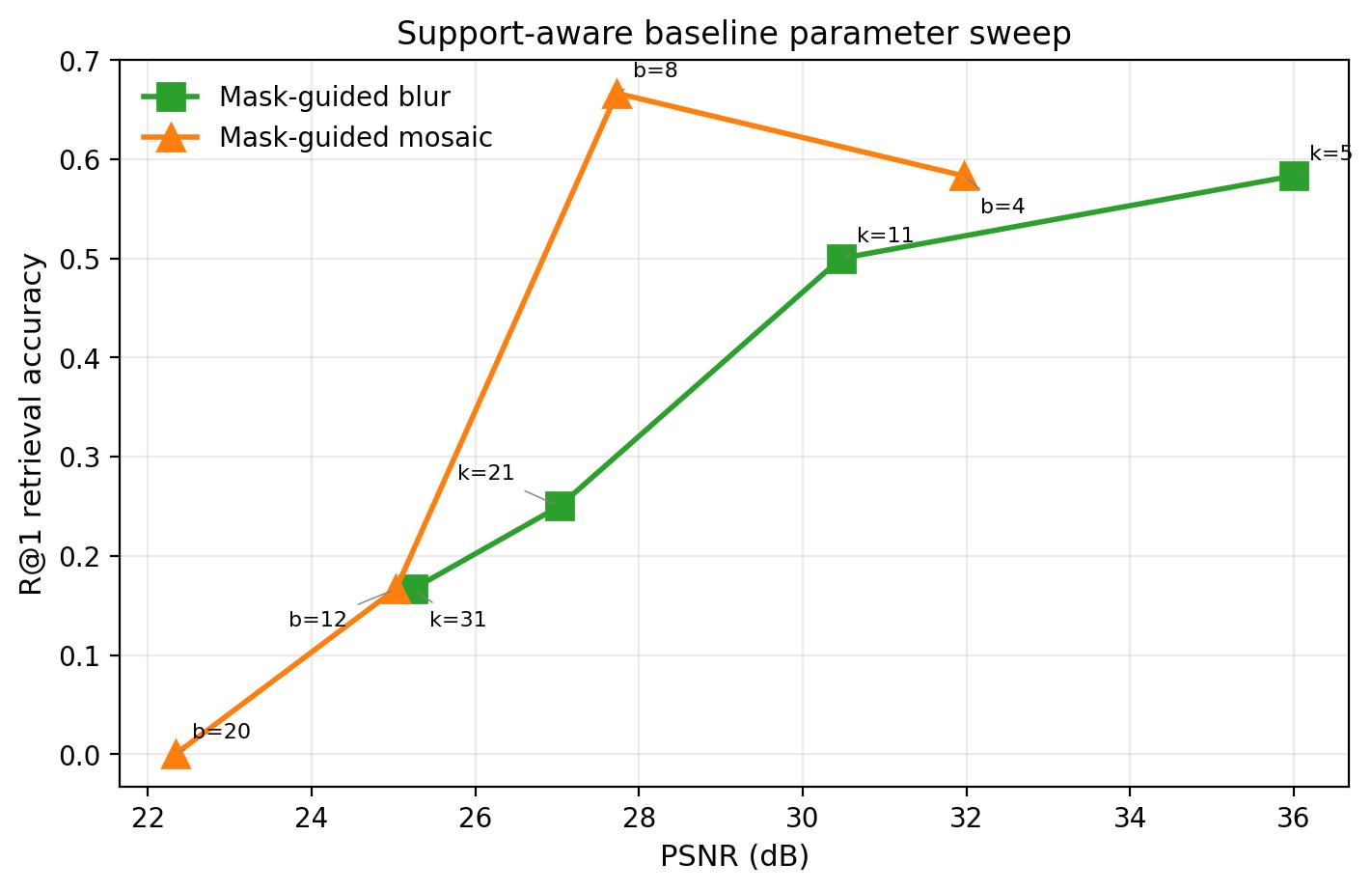}
    \caption{Support-aware deterministic baseline sweep. Each point shows a blur kernel size ($k$) or mosaic block size ($b$) applied on the same DCRF support. Blur traces a monotonically decreasing privacy--utility curve; mosaic is non-monotonic, with intermediate block sizes preserving retrieval while larger blocks eliminate it.}
    \Description{A two-curve privacy-utility plot for support-aware deterministic baselines, with blur and mosaic parameter sweeps showing R@1 retrieval accuracy versus PSNR.}
    \label{fig:baseline_sweep}
\end{figure}

These analyses clarify the practical value of PPEDCRF. Its primary advantage is in the \emph{high-quality operating region}: at any given noise budget, selective calibration preserves $\approx$6\,dB more PSNR than indiscriminate noise, which matters most when released frames must remain visually useful for downstream analytics. When the utility budget is reduced, the stochastic methods converge in privacy, and the deterministic support-aware baselines provide a complementary operating point---confirming that support localization is the most consistently beneficial component in the current evaluation. To make this comparison explicit, Fig.~\ref{fig:baseline_sweep} visualizes the blur-kernel and mosaic-block parameter sweep on the same DCRF support. Once the DCRF support is fixed, blur and mosaic trace distinct privacy--utility behaviors on that support, confirming that the support itself is the most consistently beneficial component while the perturbation rule determines the operating curve laid on top of it.

\subsection{Temporal Consistency and Perturbation Stability}
\label{sec:temporal}

Because PPEDCRF processes video sequences, temporal consistency of the perturbation is important for both perceptual quality and defense robustness. We measure two temporal metrics across all variants, summarized in Table~\ref{tab:temporal}.

\textbf{Flicker score.}
The mean absolute frame-to-frame difference in the sanitized output (lower indicates smoother temporal behavior). Table~\ref{tab:temporal} shows that PPEDCRF ($4.80$) and its ablations (w/o temporal: $4.80$; w/o NCP: $4.81$) achieve comparable flicker, all substantially below Random mask ($5.97$) and Global Gaussian noise ($9.06$), which show higher flicker because their perturbation varies independently each frame. Notably, mask-guided blur ($0.93$) and mosaic ($1.10$) achieve the lowest flicker because deterministic methods produce no inter-frame noise variation.

\textbf{Perturbation stability.}
The standard deviation of per-frame perturbation energy (lower indicates more consistent protection). PPEDCRF achieves perturbation stability of $0.006\pm0.001$, comparable to its ablations. Random mask has the highest stability variance ($0.015$) due to independent per-frame mask sampling. Global Gaussian noise has intermediate variance ($0.010$) because per-frame noise self-averages across the full frame.

\begin{table}[t]
\centering
\caption{Temporal consistency metrics ($\sigma_0=8$). Flicker: mean absolute frame-to-frame difference (lower $=$ smoother). Perturbation stability: standard deviation of per-frame perturbation energy (lower $=$ more consistent). Mask IoU: temporal overlap of perturbation support (higher $=$ more stable). Deterministic baselines report exact values; stochastic variants report mean$\pm$std over three seeds.}
\label{tab:temporal}
\begin{tabular}{lccc}
\hline
Method & Flicker $\downarrow$ & Pert.~Stab.~$\downarrow$ & Mask IoU $\uparrow$ \\
\hline
PPEDCRF & $4.80\pm0.001$ & $0.006\pm0.001$ & 0.998 \\
w/o temporal & $4.80\pm0.001$ & $0.005\pm0.001$ & 1.000 \\
w/o NCP & $4.81\pm0.001$ & $0.005\pm0.001$ & 0.998 \\
Mask-guided blur & $0.93$ & $0.082$ & 0.998 \\
Mask-guided mosaic & $1.10$ & $0.088$ & 0.998 \\
Random mask & $5.97\pm0.007$ & $0.015\pm0.002$ & 0.334 \\
Global Gaussian noise & $9.06\pm0.001$ & $0.010\pm0.001$ & 1.000 \\
\hline
\end{tabular}
\end{table}

These temporal metrics complement the static ablation in Table~\ref{tab:ablation}: structured support-aware perturbation is more stable than unstructured alternatives such as random masks and global Gaussian noise. Within the structured methods, deterministic baselines remain the temporally smoothest, while the difference between full PPEDCRF and its temporal ablation is negligible on these short clips, suggesting that the current benchmark is not highly sensitive to the temporal component itself.

\subsection{Qualitative Inspection}
\begin{figure*}[!t]
    \centering
    \includegraphics[width=\textwidth]{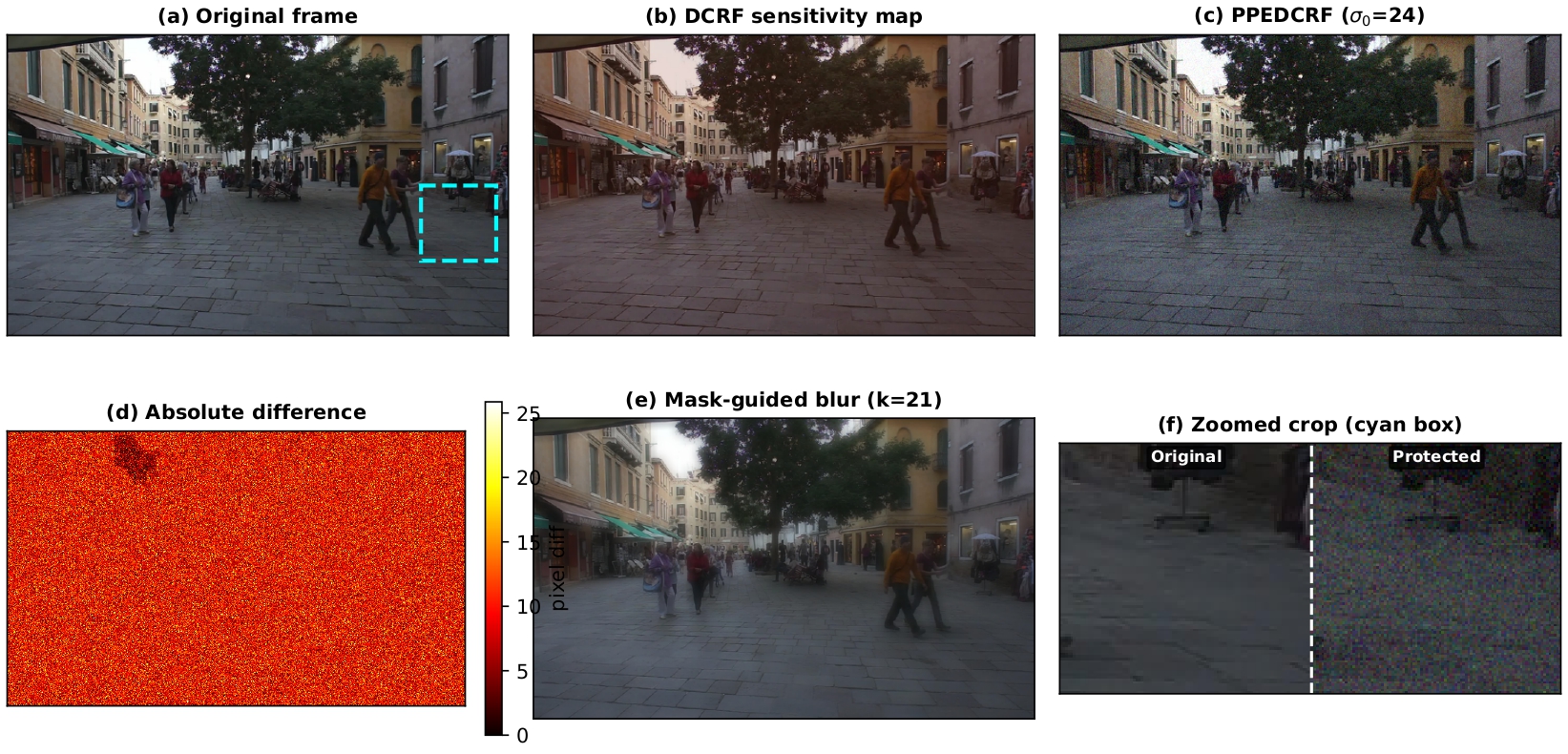}
    \caption{Qualitative visualization of PPEDCRF sanitization ($\sigma_0=24$ for visual clarity; at lower $\sigma_0$ the difference map shows sharper spatial selectivity). (a)~Original video frame. (b)~DCRF sensitivity heatmap overlay, where warm colors indicate high location sensitivity in the background. (c)~Noise-protected output. (d)~Absolute pixel difference showing spatially selective perturbation. (e)~Mask-guided blur on the same support. (f)~Zoomed crop comparing original (left) and protected (right) in a high-sensitivity region.}
    \Description{Six-panel qualitative visualization: original frame, DCRF sensitivity heatmap overlay, noise-protected frame, pixel difference map, mask-guided blur result, and zoomed side-by-side crop of a high-sensitivity background region.}
    \label{fig:sidebyside}
\end{figure*}

Figure~\ref{fig:sidebyside} provides a detailed visual inspection of the release-side mechanism. Panel~(b) overlays the DCRF-inferred sensitivity map on the original frame, highlighting which background regions are identified as location-revealing. The noise-protected output in~(c) shows visible perturbation concentrated on these high-sensitivity regions, while the pixel-wise difference map in~(d) confirms that the perturbation energy is spatially localized rather than uniformly spread. Panel~(e) shows the alternative mask-guided blur on the same support, which smooths rather than randomizes the background. The zoomed crop in~(f) makes the perturbation intensity directly comparable between the original and protected versions. This qualitative behavior is consistent with the quantitative frontier in Fig.~\ref{fig:tradeoff}: PPEDCRF changes less of the frame than the global Gaussian baseline, preserving more visual content while targeting location-relevant background structures.

\section{Conclusion}
This paper presents PPEDCRF, a calibrated selective perturbation framework for \emph{background-based location privacy} in released video frames. PPEDCRF concentrates perturbation energy on DCRF-inferred location-sensitive background regions, modulates per-pixel strength via a normalized control penalty, and uses DP-style Gaussian calibration to provide a continuously adjustable noise budget.

Three findings emerge from the paired-scene retrieval benchmark. First, selective perturbation preserves approximately 6\,dB more PSNR than indiscriminate global Gaussian noise at comparable privacy reduction---a consistent advantage across $\sigma_0\in\{8,16,24\}$. Second, the matched-operating-point analysis confirms that at matched utility around $\approx$30\,dB PSNR, PPEDCRF and global Gaussian noise converge to the same Top-1 privacy; PPEDCRF's practical advantage is preserving higher utility at the same nominal noise scale. Third, seed-averaged transfer across all eight attacker backbones shows consistent negative $\Delta$ in 23 of 24 backbone--gallery cells, including CLIP ViT-L/14 and dedicated VPR embedders. Seed averaging across three independent noise realizations stabilizes the transfer assessment in the main protocol, while appendix-scale results identify MixVPR as a remaining adverse-transfer exception.

These conclusions are established on a controlled paired-scene proxy benchmark with 12 pairs, eight attacker backbones, galleries of up to 48 candidates, and three noise seeds. A scaling confirmation with 50 pairs and galleries up to 100 candidates (Appendix) corroborates the main findings: 21 of 24 cells show negative~$\Delta$ and the overall pattern matches the 12-pair benchmark, while MixVPR remains a consistent adverse-transfer exception.

The main practical value of PPEDCRF is not universal attacker-agnostic privacy, but a calibrated release-side mechanism that offers favorable privacy--utility operating points when combined with transferable support localization. Remaining limitations include the proxy nature of the paired-scene benchmark (mined by visual similarity rather than GPS ground truth) and the absence of large cross-view or street-view evaluation. Appendix~A provides implementation-alignment notes and a 50-pair scaling confirmation that tests whether the main 12-pair conclusions persist under a larger evaluation set. Future work should address these gaps while also studying attacker-adaptive perturbation strategies beyond fixed Gaussian calibration.

\ifIEEEFORMAT
\bibliographystyle{IEEEtran}
\else
\bibliographystyle{ACM-Reference-Format}
\fi
\bibliography{ref}

\end{document}


\maketitle

\section*{Appendix}

\section{Released Implementation Notes}
The released code path used in this revision is intentionally lightweight and implementation-aligned. The unary module predicts a single-channel sensitivity logit map, the dynamic CRF refines that map with average-pooling-based spatial smoothing and previous-frame reuse, and the noise injector perturbs the \emph{released image} rather than a hidden feature tensor. This point matters because earlier drafts described feature-space release and image reconstruction more aggressively than the implementation justified.

The normalized control penalty (NCP) is also implemented conservatively in the released code. Rather than building a large symbolic hierarchy at runtime, the current implementation normalizes the refined sensitivity map and uses it as a per-pixel perturbation controller. The released sanitizer intentionally multiplies the refined support map and the normalized control map, so high-confidence sensitive pixels receive the strongest perturbation inside the same spatial support. The manuscript therefore treats the hierarchy-based interpretation as motivation, while the empirical sections report results only for the released normalization rule.

\section{Controlled Paired-Scene Retrieval Benchmark}
The revised manuscript adds a controlled paired-scene retrieval benchmark implemented in \texttt{run\_controlled\_retrieval\_benchmark.py} under \texttt{src/scripts/}. The benchmark construction is deterministic and follows four steps:
\begin{enumerate}
    \item Index monitoring sequences whose frame names follow the pattern \texttt{<clip\_id>\_frame<number>.jpg}.
    \item Extract one representative frame per sequence and embed it with a fixed ResNet18 backbone.
    \item Greedily pair the most similar sequences into synthetic location pairs, using one sequence for gallery views and the paired sequence for query views.
    \item Add hard distractors by selecting unused sequences that are maximally similar to the paired locations.
\end{enumerate}

For the final reported setting, the script constructs 12 paired locations and 36 hard distractors, which yield gallery sizes 12, 24, and 48. Query views are sanitized with three random noise seeds (1234, 1235, 1236), but retrieval is evaluated on the sanitized middle frame of each short query sequence; the surrounding frames provide temporal context only for the mask inference stage. All eight attacker backbones share the same pair discovery and gallery construction. The script exports both paper-facing figures and CSV summaries under \texttt{src/outputs/controlled\_retrieval\_seed\_avg/}.

The released selection artifact also records benchmark hardness statistics. In the current paper setting, the cosine similarity of the retained gallery--query pairs under the anchor ResNet18 embedder averages 0.989 (minimum 0.980, maximum 0.997). The selected hard distractors are not random negatives: the first 12 distractors used for the 24-way gallery average 0.919 maximum similarity to one of the paired locations, while all 36 distractors used for the 48-way gallery average 0.856 and the hardest distractor reaches 0.994. These values explain why Top-1 retrieval remains non-trivial even before protection.

The benchmark is evaluated at inference time with fixed retrieval backbones. The raw-query baseline is deterministic for a fixed gallery, while sanitized variants are averaged over the three noise seeds. All eight backbones (ResNet18, ResNet50, VGG16, CLIP ViT-B/32, CLIP ViT-L/14, CosPlace, MixVPR, Patch-NetVLAD) are evaluated in a single unified run, so all tables in the main paper share the same pair discovery and gallery construction. In the 48-candidate default setting, Top-10 retrieval saturates at 1.0 for most variants, so the main paper focuses on Top-1 and Top-5 while preserving the full Top-10 outputs in the released CSV files.

\section{Interpretation of the Added Benchmark}
The controlled benchmark is designed to isolate the release-side privacy mechanism when a full geo-localization dataset is not locally available. It should therefore be read as a \emph{paired-scene proxy benchmark}, not as a replacement for large-scale street-view localization evaluation. Its strengths are reproducibility, controlled hard negatives, and direct attack-side measurement. Its limitations are that the sequences are shorter and less dynamic than real driving videos, and the paired locations are mined by visual similarity rather than by GPS ground truth.

This interpretation is important for the revised manuscript. Results from the benchmark support claims about privacy--utility trade-offs, masking strategy, and attacker sensitivity. They do not by themselves prove universal robustness against every retrieval backbone, and they are not presented as a formal differential-privacy certificate.

\section{Scaling Confirmation: 50 Paired Locations}
To verify that the conclusions from the 12-pair main benchmark generalize to a larger evaluation set, we run a scaling confirmation with 50 paired locations, gallery sizes 50/75/100, eight attacker backbones, three noise seeds, and 128 external COCO distractors supplementing the monitoring-only negatives.

Table~\ref{tab:large50} reports Top-1 retrieval accuracy for each backbone--gallery combination. Across all 24 cells, 21 show negative $\Delta$ (privacy improvement) and 3 are positive---all from MixVPR, which consistently exhibits adverse transfer in this larger setting. The overall pattern closely matches the 12-pair benchmark: ResNet18 shows the strongest improvement ($\Delta \approx -0.36$), CLIP ViT-L/14 remains consistently negative, and CosPlace shows small but stable gains. This confirms that the 12-pair results are not an artifact of benchmark scale.

\begin{table*}[t]
\centering
\caption{Scaling confirmation: 50 paired locations, galleries 50/75/100, $\sigma_0=8$, three seeds. $\Delta$ = PPEDCRF $-$ raw; negative values indicate privacy improvement.}
\label{tab:large50}
\begin{tabular}{lccccc}
\hline
Backbone & Gallery & Raw Top-1 $\downarrow$ & PPEDCRF Top-1 $\downarrow$ & $\Delta$ \\
\hline
ResNet18 & 50 & 0.620 & $0.260\pm0.000$ & $-0.360$ \\
ResNet18 & 75 & 0.600 & $0.240\pm0.000$ & $-0.360$ \\
ResNet18 & 100 & 0.600 & $0.240\pm0.000$ & $-0.360$ \\
\hline
ResNet50 & 50 & 0.240 & $0.133\pm0.042$ & $-0.107$ \\
ResNet50 & 75 & 0.200 & $0.087\pm0.031$ & $-0.113$ \\
ResNet50 & 100 & 0.200 & $0.087\pm0.031$ & $-0.113$ \\
\hline
VGG16 & 50 & 0.320 & $0.187\pm0.012$ & $-0.133$ \\
VGG16 & 75 & 0.240 & $0.140\pm0.020$ & $-0.100$ \\
VGG16 & 100 & 0.240 & $0.140\pm0.020$ & $-0.100$ \\
\hline
CLIP ViT-B/32 & 50 & 0.240 & $0.133\pm0.012$ & $-0.107$ \\
CLIP ViT-B/32 & 75 & 0.220 & $0.073\pm0.012$ & $-0.147$ \\
CLIP ViT-B/32 & 100 & 0.220 & $0.073\pm0.012$ & $-0.147$ \\
\hline
CLIP ViT-L/14 & 50 & 0.220 & $0.093\pm0.031$ & $-0.127$ \\
CLIP ViT-L/14 & 75 & 0.160 & $0.093\pm0.031$ & $-0.067$ \\
CLIP ViT-L/14 & 100 & 0.160 & $0.093\pm0.031$ & $-0.067$ \\
\hline
CosPlace & 50 & 0.280 & $0.247\pm0.012$ & $-0.033$ \\
CosPlace & 75 & 0.240 & $0.227\pm0.012$ & $-0.013$ \\
CosPlace & 100 & 0.240 & $0.227\pm0.012$ & $-0.013$ \\
\hline
MixVPR & 50 & 0.160 & $0.207\pm0.042$ & $+0.047$ \\
MixVPR & 75 & 0.080 & $0.160\pm0.053$ & $+0.080$ \\
MixVPR & 100 & 0.080 & $0.160\pm0.053$ & $+0.080$ \\
\hline
Patch-NetVLAD & 50 & 0.420 & $0.307\pm0.012$ & $-0.113$ \\
Patch-NetVLAD & 75 & 0.340 & $0.220\pm0.020$ & $-0.120$ \\
Patch-NetVLAD & 100 & 0.340 & $0.220\pm0.020$ & $-0.120$ \\
\hline
\end{tabular}
\end{table*}

The 50-pair benchmark also reports quality metrics comparable to the 12-pair setting: PPEDCRF achieves 36.13\,dB PSNR and 0.916 SSIM, while global Gaussian noise reaches 30.16\,dB---maintaining the $\approx$6\,dB selective-perturbation advantage. MixVPR remains the sole backbone with consistent adverse transfer, reinforcing the interpretation that PPEDCRF's privacy effect is bounded and backbone-dependent rather than universal.

\section{Legacy Detector and Segmentation Utility Track}
The original PPEDCRF repository contains a legacy training-time utility track on MOT16/17, Cityscapes, KITTI, and VOC2008. These results are preserved in this appendix as auxiliary evidence only. They indicate that selective perturbation can preserve core downstream perception behavior, but they do not directly evaluate location retrieval privacy under a fixed attacker.

\begin{figure*}[!t]
    \centering
    \includegraphics[width=\textwidth]{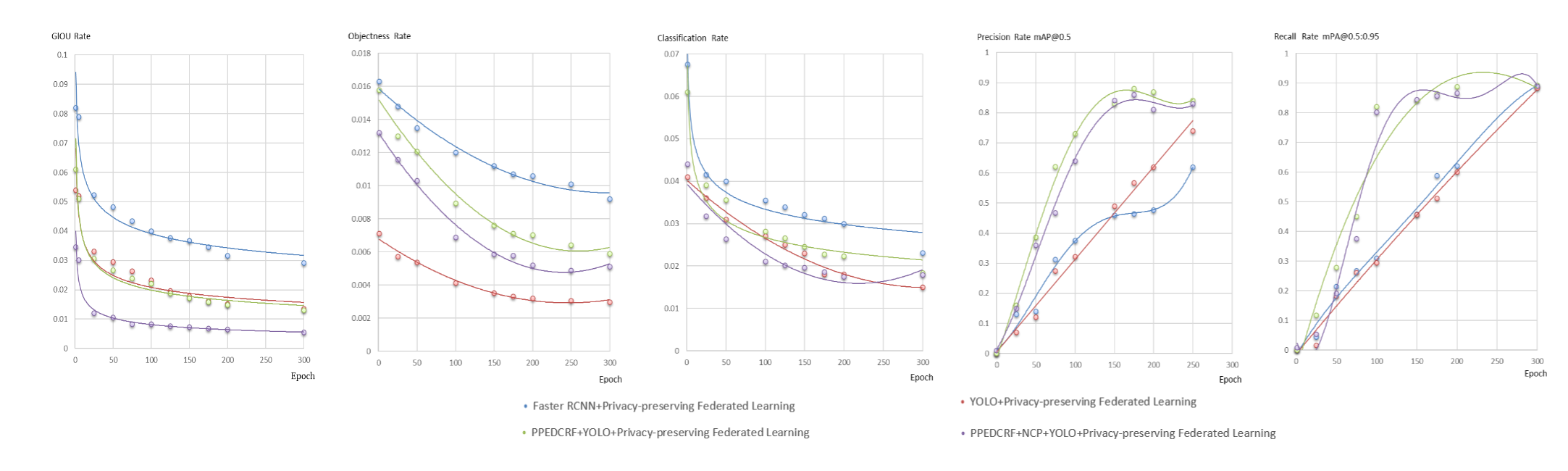}
    \caption{Legacy detector-training curves from the original PPEDCRF pipeline. These curves are retained as auxiliary downstream-utility evidence and are not the primary proof of location privacy.}
    \Description{A grid of legacy detector-training curves comparing several privacy settings over training epochs, included as auxiliary downstream-utility evidence.}
    \label{fig:app_obpwr4}
\end{figure*}

\begin{figure*}[!t]
    \centering
    \includegraphics[width=\textwidth]{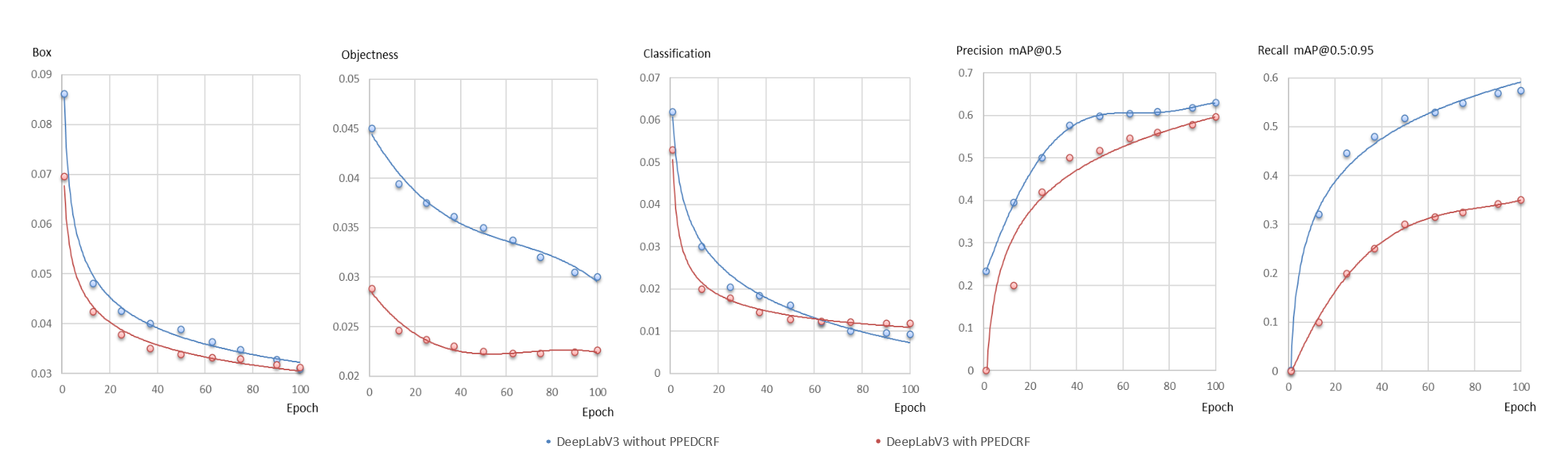}
    \caption{Qualitative utility examples from the legacy segmentation pipeline. Sanitization changes background appearance while preserving most foreground readability.}
    \Description{Several qualitative examples comparing original and sanitized frames, showing that background appearance changes while foreground objects remain readable.}
    \label{fig:app_obpwr1a}
\end{figure*}

Figure~\ref{fig:app_obpwr4} summarizes detector-training curves from the legacy pipeline. PPEDCRF-compatible training remains close to the non-private detector on objectness and precision/recall trends, while the NCP-enhanced variant remains competitive on GIoU and mAP. Figure~\ref{fig:app_obpwr1a} shows qualitative segmentation examples where sanitization changes background appearance while keeping foreground and scene structure readable.

We intentionally interpret these legacy outputs conservatively: they provide context on downstream utility, not proof of privacy robustness. The primary privacy conclusions in the revised manuscript are therefore based on the controlled paired-scene retrieval benchmark in the main text.

